\pdfoutput=1

\documentclass[11pt]{article}
\usepackage{acl}
\usepackage{times}
\usepackage{graphicx}
\usepackage{latexsym}
\usepackage{CJKutf8}
\usepackage{graphicx}
\usepackage{colortbl}
\usepackage{times}

\usepackage{lipsum}
\usepackage{multicol}

\usepackage{tabularx}
\usepackage{booktabs}
\usepackage{amssymb}
\usepackage{amsmath}
\usepackage{float}
\usepackage[utf8]{inputenc}

\usepackage{microtype}
\usepackage{graphicx}

\usepackage{graphicx}
\usepackage{latexsym}
\usepackage{CJKutf8}
\usepackage{graphicx}
\usepackage{colortbl}
\usepackage{changepage}

\usepackage{xcolor}
\usepackage{tcolorbox}
\usepackage{enumitem}  

\usepackage{mathrsfs}
\usepackage{graphicx}
\usepackage{array}
\usepackage{booktabs}
\usepackage{tabularx}
\usepackage{multirow}
\usepackage{verbatim}
\usepackage{colortbl}
\usepackage{arydshln}

\usepackage{amsfonts,amssymb}
\usepackage{bm}
\usepackage{array}
\usepackage{booktabs}
\usepackage{tabularx}
\usepackage{multirow}
\usepackage{verbatim}
\usepackage{colortbl}
\usepackage{arydshln}
\usepackage{amssymb}
\usepackage{amsmath} 
\usepackage[normalem]{ulem}
\usepackage{multirow}
\usepackage{rotating}
\usepackage{caption}
\usepackage{subcaption}
\usepackage{tabularx}
\usepackage{booktabs}
\usepackage{supertabular}
\usepackage{dsfont}
\usepackage{geometry}
\usepackage{array}
\usepackage{xcolor}
\usepackage{enumitem}

\usepackage[OT2, T1]{fontenc}
\usepackage[russian, english]{babel}
\usepackage{CJKutf8}

\definecolor{pastelgreen}{HTML}{97D077}
\definecolor{customisedblue}{HTML}{6C8EBF}
\definecolor{lightblue}{HTML}{DAE8FC}

\usepackage{tcolorbox}

\usepackage{mathrsfs}
\usepackage{graphicx}
\usepackage{array}
\usepackage{booktabs}
\usepackage{tabularx}
\usepackage{multirow}
\usepackage{verbatim}
\usepackage{colortbl}
\usepackage{arydshln}
\usepackage{amsfonts,amssymb}
\usepackage{bm}
\usepackage{array}
\usepackage{booktabs}
\usepackage{tabularx}
\usepackage{multirow}
\usepackage{verbatim}
\usepackage{colortbl}
\usepackage{arydshln}
\usepackage{amsmath} 
\usepackage[normalem]{ulem}
\usepackage{multirow}
\usepackage{rotating}
\usepackage{caption}
\usepackage{subcaption}
\usepackage{tabularx}
\usepackage{booktabs}
\usepackage{supertabular}
\usepackage{dsfont}

\usepackage[utf8]{inputenc} % Allow UTF-8 input
\usepackage[T1]{fontenc}    % Use 8-bit T1 fonts

\usepackage{url}
\usepackage{array}
\usepackage{xspace}
\usepackage{verbatim}
\usepackage{float}
\usepackage{ifthen}
\usepackage{multicol}
\usepackage{amsmath}
\usepackage{booktabs}       % Professional-quality tables
\usepackage{makecell}
\usepackage{multirow}
\usepackage{tabu}
\usepackage{threeparttable}
\usepackage{tabularx}
\usepackage{tabulary}
\usepackage{amsfonts}       % Blackboard math symbols
\usepackage{amsmath}
\usepackage{mathtools}
\usepackage{nicefrac}       % Compact symbols for 1/2, etc.

\usepackage{algorithm}
\usepackage{algpseudocode}

\usepackage{tikz}
\usetikzlibrary{bayesnet, shapes, arrows, positioning, arrows.meta}
\usepackage{pgfplots}
\usepackage{pgfplotstable}
\pgfplotsset{compat=newest}
\usepackage{hyperref}
\usepackage{cleveref}

\usepackage[frozencache,cachedir=minted-cache]{minted}
\usepackage{listings}

\usepackage[normalem]{ulem}
\usepackage{wrapfig}
\usepackage[compatibility=false]{caption}
\usepackage[labelformat=simple]{subcaption}

\usepackage{tcolorbox}
\usepackage{colortbl}

\usepackage{tabularx}
\usepackage{tcolorbox}
\tcbuselibrary{skins, xparse}

\usepackage{times}
\usepackage{latexsym}
\usepackage[T1]{fontenc}
\usepackage[utf8]{inputenc}
\usepackage{microtype}
\usepackage{times}
\usepackage{graphicx}
\usepackage{latexsym}
\usepackage{CJKutf8}
\usepackage{graphicx}
\usepackage{colortbl}
\usepackage{booktabs}
\usepackage{tabularx}
\usepackage{amsmath} 
\usepackage[normalem]{ulem}
\usepackage{multirow}
\usepackage{rotating}
\usepackage{caption}
\usepackage{subcaption}
\usepackage{tabularx}
\usepackage{booktabs}
\usepackage{supertabular}
\usepackage{subcaption}
\usepackage{arydshln} 
\usepackage{xspace}

\definecolor{citecolor}{RGB}{34,139,34}
\definecolor{lightred}{RGB}{241,140,142}
\definecolor{amber(sae/ece)}{rgb}{1.0, 0.49, 0.0}
\definecolor{battleshipgrey}{rgb}{0.52, 0.52, 0.51}
\definecolor{cadmiumorange}{rgb}{0.93, 0.53, 0.18}
\definecolor{applegreen}{rgb}{0.55, 0.71, 0.0}
\definecolor{cadmiumgreen}{rgb}{0.0, 0.42, 0.24}
\definecolor{forestgreen}{rgb}{0.13, 0.55, 0.13}
\definecolor{red}{rgb}{0.89, 0.0, 0.13}

\newcommand{\DRAG}{\textsc{\textit{D}-RAG}\xspace}

\title{Improving Multilingual Retrieval-Augmented Language Models through Dialectic Reasoning Argumentations}

\author{
  Leonardo Ranaldi $^{(\dagger)}$  Federico Ranaldi $^{(\ddagger)}$ \\
  \textbf{Fabio Massimo Zanzotto} $^{(\ddagger)}$ \textbf{Barry Haddow}$^{(\dagger)}$ \textbf{Alexandra Birch}$^{(\dagger)}$\\
	${(\dagger)}$ School of Informatics, University of Edinburgh, UK \\ ${(\ddagger)}$ Department of Enterprise Engineering, University of Rome Tor Vergata, Italy  \\
  \texttt{\{first\_name.last\_name\}@ed.ac.uk}
}

\begin{document}
\maketitle
\begin{abstract}
Retrieval-augmented generation (RAG) is key to enhancing large language models (LLMs) to systematically access richer factual knowledge. Yet, using RAG brings intrinsic challenges, as LLMs must deal with potentially conflicting knowledge, especially in multilingual retrieval, where the heterogeneity of knowledge retrieved may deliver different outlooks.

To make RAG more analytical, critical and grounded, we introduce \textit{Dialectic-RAG} (\DRAG), a modular approach guided by \textit{Argumentative Explanations}, i.e., structured reasoning process that systematically evaluates retrieved information by comparing, contrasting, and resolving conflicting perspectives. Given a query and a set of multilingual related documents, \DRAG selects and exemplifies relevant knowledge for delivering dialectic explanations that, by critically weighing opposing arguments and filtering extraneous content, clearly determine the final response. Through a series of in-depth experiments, we show the impact of our framework both as an in-context learning strategy and for constructing demonstrations to instruct smaller models. The final results demonstrate that \DRAG significantly improves RAG approaches, requiring low-impact computational effort and providing robustness to knowledge perturbations.
\end{abstract}

\section{Introduction}
\label{sec:intro}

\begin{figure*}[h]
\centering
    \includegraphics[width=0.96\textwidth]{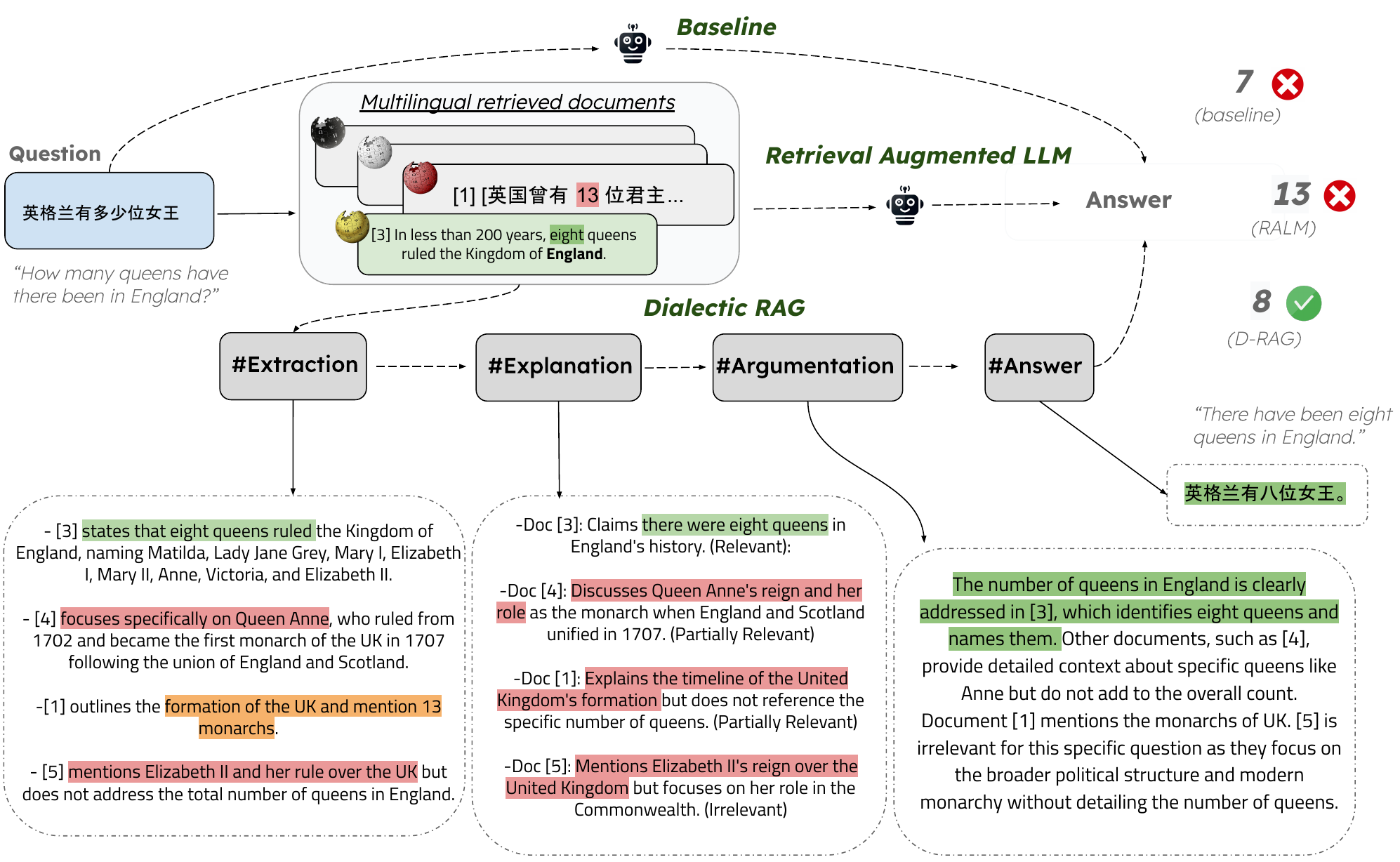}
    \caption{Our \DRAG allows LLMs to leverage multilingual knowledge-intensive question answering tasks by delivering argumentative explanations that support the final answer.}
    \label{fig:overall_pipeline}
\end{figure*}

Retrieval-augmented Generation (RAG) has emerged as a
promising approach for grounding (LLMs) responses by incorporating relevant knowledge from external sources through structured retrieval mechanisms \cite{guu2020realmretrievalaugmentedlanguagemodel}.
RAG was conceived to handle the limitations of LLMs, such as their inclination towards hallucinations and the lack of knowledge of the specialized domain in their training data \cite{siriwardhana-etal-2023-improving,zhang2023sirenssongaiocean}. 

Contextualising questions by adding relevant in-context knowledge retrieved from external corpora, such as Wikipedia, effectively reduced inaccurate generation, thereby notably improving accuracies. %\cite{gao2024retrievalaugmentedgenerationlargelanguage}. 

However, there are still limitations associated with RAGs; recent studies have shown ongoing challenges arising from the retrieved knowledge, where irrelevant or contradictory documents may introduce biases in the models \cite{menick2022teachinglanguagemodelssupport}. These weaknesses arise from the inability of RAG strategies to critically asses the retrieved knowledge \cite{ranaldi2024elicitingcriticalreasoningretrievalaugmented}. 

Prior approaches improve the RAG pipeline by incorporating external tools \cite{li-etal-2023-large,yoran2024makingretrievalaugmentedlanguagemodels} or employ multi-step reasoning strategies \cite{zhao2024empiricalstudyretrievalaugmented,zhang2024raftadaptinglanguagemodel} to determine the relevance of in-context passages. However, these methods may require high computational costs and definitely do not impact smaller-scale LLMs.  
Recently, \citet{xia2024improvingretrievalaugmentedlanguage,ranaldi2024elicitingcriticalreasoningretrievalaugmented} proposed efficient approaches to enable LLMs to deliver argumentative reasoning trajectories. Yet, their effort is on English-centric RAG, and this can be a limitation for the limited variance of retrieved knowledge and the operability \cite{chirkova2024retrievalaugmentedgenerationmultilingualsettings,ranaldi2025multilingualretrievalaugmentedgenerationknowledgeintensive}.

In this paper, we present \emph{Dialectic-RAG} (\DRAG), a modular framework conceived to enhance multilingual retrieval-augmented language models to follow a \emph{Dialectic Reasoning}, i.e., a structured analytical process that critically examines retrieved knowledge, resolves conflicting perspectives or irrelevant passages, and constructs well-supported responses through structured argumentation (Figure \ref{fig:overall_pipeline}).  To achieve this, \DRAG, employs \textit{Argumentative Explanations}, which systematically contrast opposing aspects or filter out irrelevant information irrelevant, ensuring a coherent and well-grounded final answer.
\DRAG is designed to enhance the original RAG pipeline by leading the model to leverage knowledge-intensive questions and retrieve supporting evidence through step-wise reasoning explanations that, starting from a given query, follow these steps: \textit{(a)} \textit{extraction}, where a multilingual query and documents are analysed to identify information relevant for answering the query; \textit{(b)} \textit{explanation}, where the LLMs construct single arguments about the relevance of the extracted passages, highlighting and distinguishing the furnished information; \textit{(c)} \textit{dialectic argumentation}, where the arguments are consolidated using a neutral perspective into a single final explanation; and \textit{(d)} \textit{answer}, where a short-form answer is delivered.

To evaluate the efficacy of \DRAG, we operate in two different configurations -- as an in-context approach to provide explicit instructions for larger and more capable LLMs and as a strategy for constructing synthetic demonstrations to improve the performance and align the reasoning capabilities of smaller LLMs.

Our empirical analysis carried out three different public knowledge-intensive question-answering tasks that covered 11 different languages, showing the following results and conclusions: 

\begin{itemize}
\item \DRAG elicits LLMs to deliver dialectic reasoning trajectories by exploiting multilingual knowledge in the retrieved documents, significantly enhancing performance over baselines, leading to an average accuracy increase of 51.6\% without RAG and of 12.9\% over RAG when used with GPT-4o.

\item Using \DRAG to construct synthetic dialectic multilingual reasoning demonstrations significantly improves the performance of smaller models, leading to an average increase in accuracy of 9.6\% over RAG and 5.5\% over instruction-tuning strategies for RAG when used with Llama3-8B.

\item We conduct an in-depth analysis of the components of \DRAG, showing the benefits of the components and the effects they have with definitely \textbf{contradictory} scenarios (best exemplified by the real uses-case questions presented in \textsc{BorderLines} \cite{li-etal-2024-land} augmented with multilingual retrieved documents).

\item Finally, we show that \DRAG is robust to perturbations that are a limitation for traditional RAG models, including misleading retrieval and misleading reranking (i.e., random shuffling of the retrieved documents).

\end{itemize}

\clearpage

\section{\DRAG Dialectic Reasoning in Multilingual 
 RAG}
\label{sec:methods}
Retrieval-augmented generation (RAG) enriches data access in large language models (LLMs), but they struggle to critically evaluate retrieved knowledge, handle conflicts, and filter out irrelevant content. Integrating critical reasoning into LLMs is essential to resolve information disputes and ensure more coherent and grounded responses \cite{xia2024improvingretrievalaugmentedlanguage,ranaldi2024elicitingcriticalreasoningretrievalaugmented}. 
To instruct a LLM in deliver dialectic multilingual reasoning trajectories in a Retrieval-Augmented Language Model (RAG) setting, we propose a modular strategy (Figure \ref{fig:overall_pipeline}) formed of: \textit{(a) extraction} (\S \ref{sec:Extracting}), where, given a query a set of multilingual retrieved documents, the model identify relevant information; \textit{(b) argumentation} (\S \ref{sec:Argumentative_Explanations}), where the model delivers argumentative motivations about the extracted information, by displaying and discerning the relevancy about the aspects; \textit{(c) dialectic argumentation} (\S \ref{sec:Dialectical_Reasoning}), where the arguments constructed in \textit{(b)} are summarised using a dialectic and neutral perspective into a single explanation; \textit{(d) answering} (\S \ref{sec:Dialectical_Reasoning}), where a final answer to the query is generated adhering to query constraints such as query-language and the compact form of the answer as reported in Appendix \ref{app:DRAG-prompt-annotation}.

We then use \DRAG in two scenarios as an in-context learning strategy and a synthetic generator for constructing demonstrations (\S \ref{sec:DRAG_Application}). For the in-context learning strategy, we use \DRAG to instruct LLMs to follow step-wise dialectic planning that improves the base RAG pipelines (\S \ref{sec:in-context}).
For the instruction-tuning, we use the synthetic demonstrations to improve smaller LLMs (\S \ref{sec:tuning}) and transfer to them the capability of leveraging the query and the retrieved knowledge for delivering a robust argumentation to reach the answer.

\subsection{Extraction}
\label{sec:Extracting}
The first step, which we define as $\alpha_1$ in the proposed pipeline, concerns extracting relevant retrieved knowledge from documents retrieved from a given knowledge base $\mathcal{K}$. Complementary to previous approaches \cite{xia2024improvingretrievalaugmentedlanguage,ranaldi2024elicitingcriticalreasoningretrievalaugmented} in this paper, we operate in a multilingual retrieval scope (where documents come from knowledge bases in multiple languages as defined in \S \ref{sec:dataset}). We operate via multilingual retriever systems provided by Cohere\footnote{https://huggingface.co/Cohere/Cohere-embed-multilingual-v3.0} as the default retriever model $\mathcal{R}$. 
Thereafter, we instruct the model to analyse the query and understand and identify the main points from the retrieved documents (i.e., "\textbf{\#Reference Evidence}") for answering the question and label this phase as "\textbf{\#Extraction}". Since we work with multilingual queries and documents, this step is crucial to aid the model in planning the reasoning.

\subsection{Explanations}
\label{sec:Argumentative_Explanations}
The second step, defined as $\alpha_2$, concerns instructing the model to discuss the extracted information and deliver argumentations.
Specifically, after identifying and extracting information from the top-$k$ documents, we prompt the model to discuss whether they are actually relevant or irrelevant to the query by clearly citing the passages and labelling this phase as "\textbf{\#Explanation}".

\subsection{Dialectic Reasoning}
\label{sec:Dialectical_Reasoning}
This step, which we define as $\alpha_3$, concerns generating a final comprehensive explanatory summary.
In particular, for $\alpha_3$, we leverage the arguments in the previous steps to deliver the final explanation that argues the motivations that support the answer using a dialectic approach, i.e. a critical approach that relies on systematic comparison to arrive at a more articulate and well-founded conclusion. Hence, we instruct the LLM to consider the generated aspects, summarise the main points into a single argumentation, and head this as “\textbf{\#Dialectic Argumentation:}”. 

\subsection{Final Answer}
The last step is defined as $\alpha_4$ and results in a short-form answer used in the final evaluation.
We instruct the model to generate the final answer in this form and in the same language as the query following the pattern "\textbf{\#Answer:}".

\subsection{\DRAG Application}
\label{sec:DRAG_Application}

\subsubsection{\DRAG as in-context Learning}
\label{sec:in-context}

We adopt \DRAG as in-context learning strategy by instructing different LLMs to answer knowledge-intensive questions by dealing with retrieved knowledge. \DRAG, in a modular way, identify the most critical information from the retrieved documents (\S \ref{sec:Extracting}), arguing the rationale supporting the selection of appropriate points to answer the query by explaining the main passages (\S \ref{sec:Argumentative_Explanations}), deliver a single argumentation that best describes the points (\S \ref{sec:Dialectical_Reasoning}); and finally, generate the final short-form answer in a strict format, to have a more detailed and strict downstream evaluation. 
Yet, although the sequence of instructions is well-structured and defined, the ability to perform sequential and complex reasoning tasks is limited to larger LLMs (such as GPT-4o, as discussed in the experiments). Hence, we transfer these capabilities to smaller models operating via \DRAG for building synthetic demonstrations as training sets.

\subsubsection{\DRAG as a Synthetic Annotation}
\label{sec:annotation_strategy}
We instruct smaller models via demonstrations produced by high-performing LLMs capable of following structured instructions. 
In contrast to the methods proposed in \cite{asai2023selfraglearningretrievegenerate}, we use a single prompt composed of a sequence of instructions in a multilingual setting. 
To filter the quality of generated demonstrations, we follow the method proposed by \citet{xia2024improvingretrievalaugmentedlanguage,ranaldi2024elicitingcriticalreasoningretrievalaugmented}, which computes the citation precision for the considered documents as a proxy for the quality of the demonstrations. However, since \DRAG employs a different annotation mechanism, our annotation pipelines firstly filter out the final correct answers through a strict, exact match; then, after the filtering (which removes more than half of the annotated demonstrations), it verifies that the provided instructions have been considered. We detail the description of annotation in Appendix \ref{app:annotation_phase_details}.

\subsection{Tuning Smaller Models}
\label{sec:tuning}
We fine-tune a Language Model $\theta$ using the annotations\footnote{we select annotations as described in \S \ref{sec:annotation_strategy}} generated via \DRAG. The annotations are augmented with demonstrations $\alpha$ using the standard language modelling objective to maximize the expected log-likelihood:

\begin{equation*}
\theta^{*} = \arg\max_{\theta} \mathbb{E}_{(Q, \alpha, Y) \sim \mathcal{D}} \left[ \log p_{\theta}(Y, \alpha \mid Q) \right]
\end{equation*}

\noindent where $\theta^{*}$ denotes the optimal model parameters, and $p_{\theta}(Y, \alpha \mid Q)$ is the joint probability of the output $Y$ and the demonstrations $\alpha$ conditioned on the query $Q$, learned from the training corpus $\mathcal{D}$ augmented with contrastive reasoning demonstrations. While $\alpha = \alpha_1 \cdot \alpha_2 \cdot \alpha_3 \cdot \alpha_4$ is the combination of the multiple reasoning steps performed by the model, "$\cdot$" is the concatenation operator, and $\alpha_i$ are the respective paths generated by the overhead processes. $Q$ is the provided query, and $Y$ is the output, including the intermediate steps and the final answer that compose the training corpus $\mathcal{D}$.

\section{Experimental Setup}
\label{sec:exp_set}

We evaluate \DRAG on five open-domain question-answering tasks (\S \ref{sec:dataset}). We perform the retrieval and evaluation phases by following standard approaches used to assess the RAG pipeline (\S \ref{sec:retriever_evaluation}) and perform the tuning phase by using the setup presented in \S \ref{sec:training_setting}. 

\subsection{Tasks \& Datasets}
\label{sec:dataset}
We use the following question-answering (QA) tasks: \textit{(i)} MLQA \cite{lewis-etal-2020-mlqa}, \textit{(ii)} MKQA \cite{longpre2021mkqalinguisticallydiversebenchmark} and \textit{(iii)} XOR-TyDi QA \cite{asai-etal-2021-xor} as they best represent multilingual open-ended question-answering tasks. Then, we use \textsc{Borderlines} \cite{li-etal-2024-land}, which contains multilingual questions concerning conflicts over disputed territories (note: we follow the questions and targets delivered by \citet{li-etal-2024-land}). Finally, we include Natural Questions (NQ) \cite{10.1162/tacl_a_00276}, as it is a widely used English benchmark for assessing RAG systems. This allows us to establish meaningful baselines for comparison.  Appendices \ref{app:info_dataset} and \ref{app:languages} report the languages and composition of each dataset. Appendix \ref{app:exps_borderlines} reports detailed information about \textsc{Borderlines}.

\subsection{Experimental Settings}
\label{sec:retriever_evaluation}

\paragraph{Retrieval}
In our work, we employ Wikipedia as the knowledge base $\mathcal{K}$ and Cohere as the retrieval system $\mathcal{R}$. Specifically, by working through the Wikimedia dump provided by Cohere\footnote{Cohere/wikipedia-2023-11-embed-multilingual-v3}, individual articles are embedded with the state-of-the-art multilingual embedding model \textit{Cohere\_Embed\_V3}. This pipeline makes it easy to search Wikipedia for information or to use only specific languages. 
For each question in the evaluation data, we retrieve the top-5 relevant documents (details Appendix \ref{app:retrieval}). 

\paragraph{Models \& Inference Settings}

To get a comprehensive evaluation of existing RAG pipelines in the main experiments, we use four different LLMs: GPT-4o \cite{openai2023gpt4}, Llama3-70b-instruct \cite{grattafiori2024llama3herdmodels} and smaller models Llama3-8b-instruct and 1b-instruct\footnote{to simplify notation we omit \textit{instruct} for the rest of the paper}. Detailed settings and model versions are in Appendix \ref{app:model_versions}.
We use greedy decoding in all experiments to ensure a more deterministic generation process, and we set the temperature to 0 and the maximum generation length to 2048. We observed that these settings deliver better and deterministic performances.

\subsection{Evaluation Metrics}
\label{sec:evaluation}
We use flexible exact-match accuracy following \citet{schick2023toolformer}, which is based on whether or not ground-truth answers are included in the generated answers provided by the models instead of a strict exact match. Moreover, our prompting pipelines instruct the models to use as a final label \textbf{‘\#Answer’} (see Appendices \ref{app:app_prompting}) to elicit a conclusive generation that contains a short-form answer.

\subsubsection{Training Setting}
\label{sec:training_setting}

To evaluate the impact of \DRAG reasoning demonstrations on smaller models (\S \ref{sec:methods}), we employ the annotations produced following the \DRAG strategy (\S \ref{sec:annotation_strategy}). Further, for a fair comparison, we deliver annotations using Llama-3-SFT, where Llama is tuned on training samples without \DRAG (annotation generated using same query, retrieved documents and the prompt in Table \ref{tab:baseline_RAG_prompt}). 
We fine-tune the models for three epochs with a batch size of 32 and a learning rate equal to 1e-5 with a 0.001 weight decay. We use the cosine learning rate scheduler with a warmup ratio of 0.03. We conducted our experiments on a workstation with four Nvidia RTX A6000 and 48GB of VRAM.

\subsection{Evaluated Methods}
\label{sec:prompting}
We propose the following settings:
\paragraph{Baseline - without RAG}
We evaluate the baseline capabilities of selected models in a zero-shot way without introducing any documents (without RAG) using the instruction (prompt) in Table \ref{tab:baseline_prompt}.

\paragraph{Retrieval Augmented LLM (RAG)}
We assess the impact of retrieved knowledge by instructing the evaluated models to consider the $top$-$5$ retrieved documents. We use the retrievers in \S \ref{sec:retriever_evaluation}.

\noindent \textbf{$\rightarrow$ ICL}
As baseline settings we use the instruction in Table \ref{tab:baseline_RAG_prompt}.

\noindent \textbf{$\rightarrow$ \DRAG (ICL)}
To complete the RAG-based settings, we use \DRAG as an in-context learning strategy as in Table \ref{tab:annotation_prompt_D-RAG}.

\noindent \textbf{$\rightarrow$ fine-tuning}
Finally, we tune Llama models using \textit{SFT} and \textit{\DRAG} as presented in \S \ref{sec:training_setting} and prompt using RAG instruction (Table \ref{tab:baseline_RAG_prompt}).

\begin{table}[h]
\small
\centering
\begin{tabular}{p{2.4cm}cccc}  
 \toprule 
 \multirow{1}{*}{\textbf{Models}} & \multicolumn{1}{c}{\textbf{MKQA}} & \multicolumn{1}{c}{\textbf{MLQA}} & \multicolumn{1}{c}{\textbf{X.TyDi}} & \multicolumn{1}{c}{\textbf{Avg}} \\  
 \midrule
\multicolumn{5}{c}{\textbf{Baseline}} \\  
 \midrule 
Llama3-1B & 32.5 & 33.7 & 27.3 & 31.2 \\ 
Llama3-8B & 38.9 & 43.4 & 34.5 & 38.6 \\  
Llama3-70B & 40.7 & 43.9 & 36.5 & 40.4 \\ 
GPT-4o & 44.8 & 46.9 & 36.7 & 42.8 \\   
\midrule
\multicolumn{5}{c}{\textbf{RAG}} \\  
 \midrule 
Llama3-1B & 50.6 & 48.6 & 41.7 & 46.9 \\ 
Llama3-8B & 57.3 & 54.5 & 48.1 & 53.1 \\  
Llama3-70B & 60.1 & 56.6 & 49.2 & 55.3 \\
GPT-4o & 61.4 & 58.6 & 51.2 & 57.4 \\  
\midrule 
\multicolumn{5}{c}{\textbf{RAG $\rightarrow$ \DRAG} as ICL} \\  
 \midrule 
Llama3-1B & 48.6 & 48.0 & 38.3 & 45.0 \\ 
Llama3-8B & 56.7 & 53.5 & 48.1 & 52.8 \\  
Llama3-70B & 67.3 & 62.4 & 55.8 & 62.4 \\
GPT-4o & \textbf{68.2} & \textbf{65.5} & \textbf{60.7} & \textbf{64.8} \\   
\midrule
\multicolumn{5}{c}{\textbf{RAG} $\rightarrow$ tuning via \textit{SFT} and  \textit{\DRAG} } \\  
 \midrule 
Llama3-1B$_{\textit{SFT}}$ & 52.1 & 50.0 & 41.3 & 47.8 \\ 
Llama3-8B$_{\textit{SFT}}$ & 60.3 & 56.3 & 48.5 & 55.0 \\ 

 \midrule 
Llama3-1B$_{\textit{\DRAG}}$ & 55.8 & 53.7 & 46.6 & 51.9 \\ 
Llama3-8B$_{\textit{\DRAG}}$ & \textbf{63.6} & \textbf{59.3} & \textbf{52.7} & \textbf{58.5} \\  
\bottomrule
\end{tabular}
\caption{Average results on multilingual QA tasks (\S \ref{sec:dataset}). Models instructed as detailed in \S \ref{sec:prompting}. In bold, best performances of ICL and fine-tuned models.}
\label{tab:results}
\end{table}

\section{Results}
\label{sec:Results}

%The empirical evaluations are reported in 
The results in Table \ref{tab:results} show that \DRAG aids the models in leveraging retrieved documents for multilingual QA tasks, showing the impact of dialectic reasoning argumentations on RAG. We found that \DRAG is effective as an in-context learning approach in larger LLMs and is notably helpful as a demonstration strategy to improve the performance of smaller models, achieving solid performance compared to fine-tuning approaches. 
To this end, the following sections analyse the impact of \DRAG when adopted as both an in-context strategy (\S \ref{sec:D-RAG_in-context}) and as a framework for generating annotations to instruct LLMs (\S \ref{sec:D-RAG_as_annotator}). Then, in \S \ref{sec:application_borderlines}, we study a practical application on \textsc{BorderLines} \cite{li-etal-2024-land}.
Finally, we investigate the role of the argumentative explanations (\S \ref{sec:impact_explainations}) and revealed evidence of robustness on challenging perturbations and functionality in low-resource settings (\S \ref{sec:ablation}).

\subsection{\DRAG in-context learning}
\label{sec:D-RAG_in-context}

Table \ref{tab:results} reports the results of \DRAG when adopted as an in-context learning (ICL) strategy for different models. We observe an overall improvement over the baseline models without retrieved documents (relative improvements of 51.4\% for GPT-4o, 54.5\% for Llama3-70B, 36.7\% for Llama3-8B and 44.2\% for Llama3-1B on absolute average score); however, the results show that the impact of \DRAG in a RAG setting emerges for GPT-4o and Llama3-70B where \DRAG achieves a general improvement of 12.9\% and 11.9\% respecting to RAG. 
In contrast, for Llama3-8B and Llama3-1B, we observe a decrease in performance compared to the RAG pipeline, suggesting that these smaller models cannot deliver the dialectic reasoning explanations required to support their responses.

\subsection{\DRAG Annotation Approach}
\label{sec:D-RAG_as_annotator}

Table \ref{tab:results} reports the impacts of \DRAG used as an annotation strategy for different smaller models (denoted as \textbf{RAG} $\rightarrow$ tuning via \textit{SFT} and  \textit{\DRAG}). \DRAG effectively enhances the performance of smaller models when employed to deliver reasoning demonstrations via GPT-4o. Hence, we found that \DRAG outperform $SFT$ approaches for both model versions. 

It emerges that both tuning strategies work well and outperform the baseline RAG approaches—for instance, Llama3-8B improves 52.8 $\rightarrow$ 55.0 average accuracy comparing RAG and $SFT$ versions. However, the models tuned via \DRAG annotations consistently surpass the $SFT$ (Llama3-1B improves +3.1 and Llama3-8B +2.9 average points). These results indicate the benefits provided by \DRAG demonstrations and their ability to efficiently elicit argumentations in smaller LMs.  Finally, we compared \DRAG with similar work focusing on English. We showed that although tuning is multilingual, \DRAG achieves sustainable performance. In contrast, related methods definitely underperform when operating in multilingual tasks.

\begin{figure}[h]
    \centering
    \includegraphics[width=0.95\linewidth]{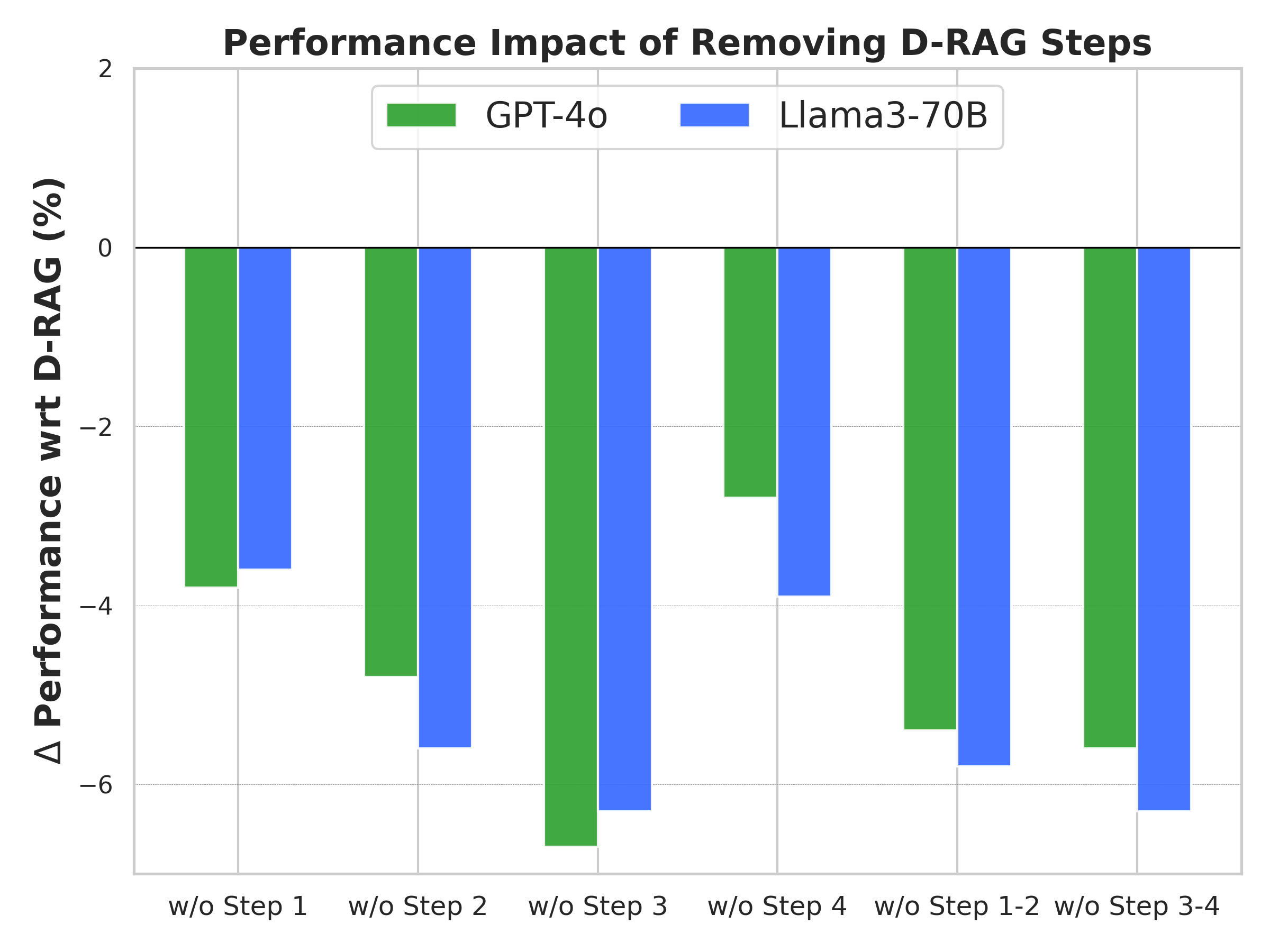}
    \caption{Performance differences $(\Delta)$ for GPT-4-o and Llama3-70B. We analyse the impact of each component on MKQA by eliminating (w/o) the \DRAG steps. }
    \label{fig:missing_steps_performances_ICL}
\end{figure}

\subsection{The Role of \DRAG Components}
\label{sec:impact_explainations}

Figures \ref{fig:missing_steps_performances_ICL} evaluate the impact of our \DRAG framework on the final performance as used as an in-context learning approach and as a tuning strategy. 
The results in Figure \ref{fig:missing_steps_performances_ICL} demonstrate the importance of each phase in the reasoning process introduced in \S \ref{sec:methods}. For GPT-4o and Llama3-70B, we observe the highest decrease in performance when removing the second and third steps. In particular, removing the second step (w/o Step 2), also defined as $\alpha_2$, which is concerned with arguing and breaking down relevant points of retrieved documents to answer the given query, it is possible to observe an average decrease of -5.2\% compared to \DRAG. Removing Step 3, which is responsible for delivering the argumentation, we observe an average reduction of -6.5\% compared to \DRAG. These results demonstrate the crucial impact of each passage of \DRAG for eliciting dialectic explanations from the model.
The impact of steps for ICL operation affects the tuning as well. As reported in detail in appendix the models tuned via modified \DRAG or randomly mixed steps negatively impact performance (the crucial points are Steps 2 and 3 as in the case of \DRAG as ICL).

\begin{figure*}[t]
\centering
         \begin{minipage}{0.32\linewidth}
     \centering
     \includegraphics[width=.95\linewidth]{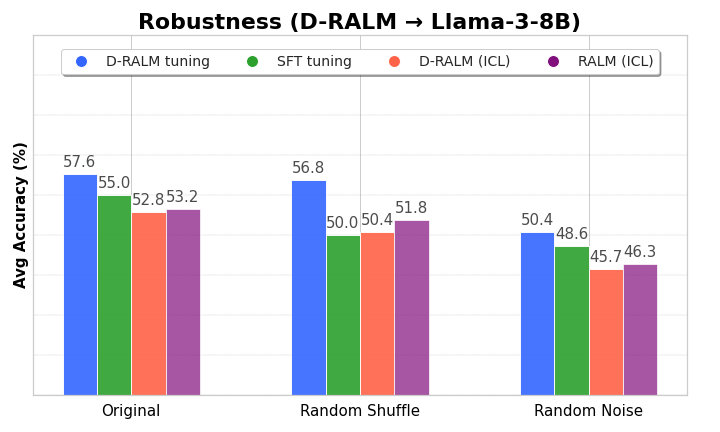}
   \end{minipage}
         \begin{minipage}{0.32\linewidth}
     \centering
     \includegraphics[width=.95\linewidth]{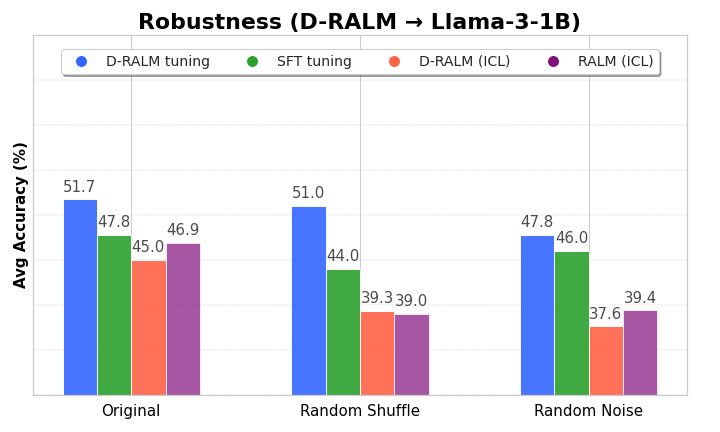}
   \end{minipage}
         \begin{minipage}{0.32\linewidth}
     \centering
     \includegraphics[width=.95\linewidth]{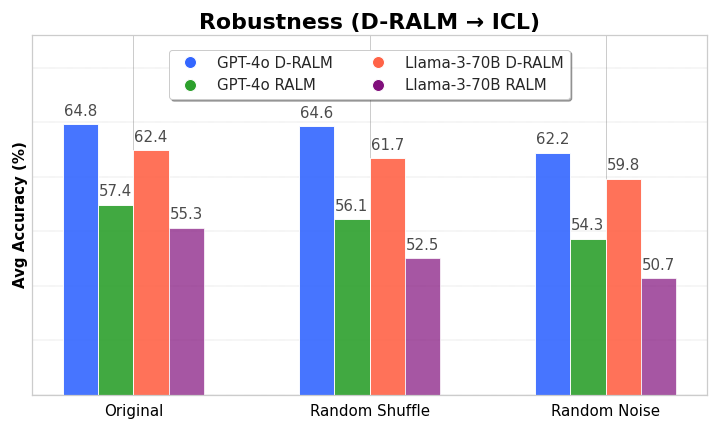}
   \end{minipage} 
   \caption{Robustness experiment results on QA datasets (\S \ref{sec:dataset}). We provide retrieved documents by randomly shuffling them (Random Shuffle) and introducing two misleading (irrelevant) documents (Random Noise).} 
   \label{fig:performances_ablation_retrieving}

\end{figure*}

\begin{table}[h]
\begin{center}
\begin{small}
\begin{tabular}{l|ccc}
\textbf{Model} & \textbf{\%Agreement}  & \textbf{\%Agreement}  \\ 
 & \textbf{English (En)} & \textbf{X,Y,En} \\

\hline
\hline
GPT-4o & 75\% &  66.6\%\\ 
+RAG & 85\%  & 81.6\% \\ 
+\DRAG & \textbf{100\%}  & \textbf{100\%} \\ 
\midrule
Llama3-8B$_{ICL}$ & 35\% &  43.3\%\\ 
+RAG$_{ICL}$ & 50\%  & 51.6\% \\ 
+\DRAG$_{ICL}$ & 65\%  & 68.3\% \\ 
\hdashline
Llama3-8B$_{\textit{SFT}}$ & 65\% &  70\%\\ 
Llama3-8B$_{\textit{\DRAG}}$ & \textbf{95\%}  & \textbf{98.3\%} \\ 
\hline
\end{tabular}
\caption{Agreement rate with controller in  \textsc{BorderLines} dataset \cite{li-etal-2024-land}. Details in Appendix \ref{app:exps_borderlines}.}
\label{tab:bolderlines_accuracies}
\end{small}
\end{center}
\end{table}

\subsection{Dialectic Reasoning in \textsc{BorderLines}}
\label{sec:application_borderlines}
To investigate the impact of our \DRAG in real contexts, we used \textsc{BorderLines} \cite{li-etal-2024-land}. This resource provides questions concerning disputed territories as detailed in Appendix \ref{app:exps_borderlines}. These questions are in English and in two additional languages, which are the land disputants (defined as \textbf{\texttt{X}} and \textbf{\texttt{Y}}). Finally, a target or controller value indicates the country that controls the territory\footnote{in some cases, there are no defined places that we do not consider in our analysis.}.
To study the consistency and dialectic capabilities of our \DRAG, we then conducted a retrieval phase and evaluated GPT-4o and Llama3-8B (tuned and not) with the questions in the specific languages and English using the prompts defined in Appendices \ref{app:app_prompting} and \ref{app:DRAG-prompt-annotation}.
Then, setting the controller as \textbf{\texttt{X}}, we estimated the percentage of times the answer provided by the models prompted in English matched with the target or named controller (denoted as \textbf{\%Agreement English}), and the percentage when the models prompted via queries in three languages matches among them and with the controller. 

Table \ref{tab:bolderlines_accuracies} shows that the consistency percentage increases when \DRAG is used. In particular, in GPT-4o, there is a 15\% and 19.6\% increase when \DRAG is compared with RAG. Similarly, it occurs between Llama3-8B instructed via \DRAG. Finally, Llama3-8B tuned with DRAG has the most robust constancy.

\subsection{Additional Analysis}
\label{sec:ablation}

\paragraph{Robustness}
To test the robustness of the proposed framework and avoid the possible performance bias obtained from noisy or misleading retrieval, we follow the methodology used in previous works. Hence, we shuffled the order of the retrieved documents \textit{(Random Shuffle)} and inserted two misleading and irrelevant documents \textit{(Random Noise)}.
Figure \ref{fig:performances_ablation_retrieving} reports the results. We show that \DRAG consistently outperforms the baseline model with RAG as ICL and annotation strategy. In particular, the random shuffling of retrieved documents minimally impacts performance, demonstrating the permutation invariance property of \DRAG (see the subfigure on the right). 
Moreover, when noisy documents are added, all the evaluated models suffer a higher performance drop. However, the drop for \DRAG is typically lower than the standard RAG approach, which shows that the proposed method is more robust even when dealing with noisier results. 

\begin{figure}[h]
    \centering
    \includegraphics[width=0.9\linewidth]{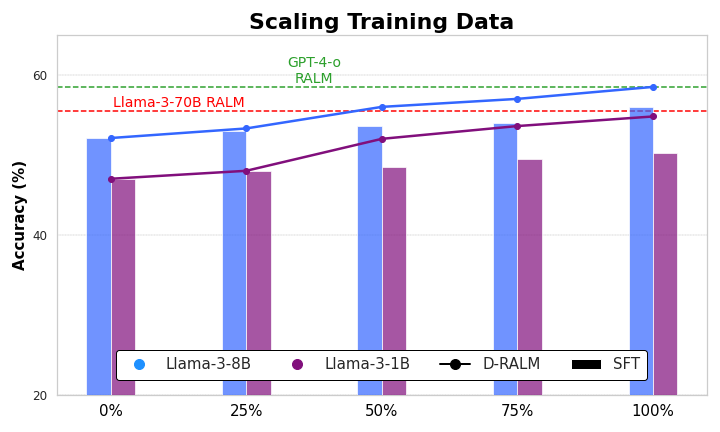}
    \caption{Performances assessment of Llama3-8B and -1B by scaling \DRAG (lines) and $SFT$ (bars) tuning demonstrations on ablation set (Appendix \ref{sec:splitting_data}).}
\label{fig:performances_scaling_demons}
\end{figure}

\paragraph{Quantity of Instructions} Figure \ref{fig:performances_scaling_demons} shows the behaviour of \DRAG when scaling-up the number of training examples. While we found that the quantity of the demonstrations used in \DRAG is important in determining the final performance, we found that \DRAG can outperform the baseline RAG models with only 50\% of training demonstrations, also achieving superior training performance when compared to the fine-tuned SFT model (i.e., the model fine-tuned without \DRAG demonstrations as explained in \S \ref{sec:exp_set}).
This further highlights the quality of the training signal provided by the contrastive explanations.

\paragraph{Quality of Generation} 
Table \ref{tab:ablation_IF_languages} shows the tendency to generate answers in the same query language and follow the provided instructions at inference time (we describe the experimental methodologies in Appendix \ref{app:ablation_outputs} ). In particular, two requirements that our framework must satisfy are \textit{i)} all instructions given in the prompt must be followed, and \textit{ii)} in the multilingual task, the answer must be in the same query language. In both cases, we observe that the GPT4-o and Llama3-70B are consistent with the requirements. On the other hand, the two Llama3 models do not follow the instructions, but when tuned, employing demonstrations from \DRAG, they become consistent. 

\begin{table}[h]
\small
\center
  \begin{tabularx}{\columnwidth}{p{2.2cm}<{\raggedright}p{0.8cm}<{\centering}p{0.8cm}<{\centering}p{1cm}<{\centering}p{1cm}<{\centering}}
    \toprule
    \multirow{1}{*}{ \textbf{Models} } & \textbf{IF} & \textbf{CL} & \textbf{\textit{LR}-IF} & \textbf{\textit{LR}-CL} \\
    \midrule
GPT-4o  & - & 85.6\% & - & 72.2\% \\
+ \DRAG$_{ICL}$  & 90.5\% & \textbf{94.8\%} & 83.6\% & \textbf{86.4\%} \\
\midrule
Llama3-70B  & - & 65.2\% & - & 63.8\% \\
+ \DRAG$_{ICL}$  & 83.5\% & \textbf{79.4\%} & 77.4\% & \textbf{70.2\%} \\
\midrule
Llama3-8B  & - & 65.9\% & - & 46.0\% \\
+ $SFT$  & - & 72.8\% & - & 64.6\% \\
\hdashline
+ \DRAG$_{ICL}$  & 58.4\% & 66.2\% & 45.5\% & 44.0\% \\
+ \DRAG$_{FT}$  & \textbf{78.3\%} & \textbf{72.0\%} & \textbf{67.1\%} & \textbf{69.6\%} \\
\midrule
Llama3-1B  & - & 57.2\% & - & 30.4\% \\
+ $SFT$  & - & 66.3\% & - & 48.8\% \\
\hdashline
+ \DRAG$_{ICL}$  & 40.0\% & 53.3\% & 40.7\% & 32.2\% \\
+ \DRAG$_{FT}$  & \textbf{60.4\%} & \textbf{69.5\%} & \textbf{45.3\%} & \textbf{59.9\%} \\
  \bottomrule
\end{tabularx}
\caption{Percentage (\%) of answers that follow the prompt instructions (IF) and generate the final answer in the correct language (CL). $FT$ indicates fine-tuned models via \DRAG. \textbf{\textit{LR}} indicates the results for low-resource languages considering the MKQA answers.}
  \label{tab:ablation_IF_languages}
\end{table}

\paragraph{\DRAG Settings \& Comparisons}

We provide evidence for the robustness of the \DRAG by proposing three experiments. Firstly, we show that decomposing our \DRAG into different prompts delivers benefits which are minimal compared to the cost of increasing the number of prompts (four prompts against a single one). Then, in Appendix \ref{app:ablation_language}, we analyse the impact of internal argumentation in the query language. As shown in Table \ref{tab:ablation_query_language}, argumentation in a language other than English (a language in which the models used in this work are more profitable) leads to a drop in performance that will definitely be a matter of future investigation. Finally, we show that \DRAG perform well even in monolingual tasks (English). In contrast, related methods achieve lower performance in multilingual tasks.

\section{Applicability \& Future Work}
Our experiments evaluate a method to improve RAG capabilities in multilingual scenarios by eliciting LLMs to consider heterogeneous sources of knowledge and argue the reasons that support the answer in a dialectic manner. The applicabilities of our work are related to: \textit{(i)} improving the answering of questions that involve a retrieval in a setting with unbalanced resource availability, e.g., in the case of Wikipedia, where the number of documents differs from languages (Table \ref{tab:language_distribution_wikimedia_dump}).
\textit{(ii)} improving the argumentation in scenarios where there is an information overlap on retrieved statements that support the outcomes, as studied in \S \ref{sec:application_borderlines}. 
\textit{(iii)} Transferring the capabilities of delivering dialectic explanations to smaller LLMs by teaching them via synthetic demonstrations.
In future developments, we plan to analyse the role different languages can play in delivering reasoning and how much the multilingual proficiency of LLMs can influence this task.

\section{Related Work}

\citet{lewis2020retrieval} investigated the advantages of augmenting LLMs with retrieved knowledge, a technique known as Retrieval-augmented Language Models (RAG). \citet{shi2023largelanguagemodelseasily} demonstrated that the benefits of RAG could be undermined by noisy retrieval. Several studies have enhanced RAG through in-context solutions, tuning, or retriever interventions \cite{menick2022teachinglanguagemodelssupport,jiang-etal-2023-active,gao-etal-2023-rarr,sawarkar2024blendedragimprovingrag}. 
While effective, in-context learning only partially mitigates retrieval bias, and tuning remains costly \cite{asai2023selfraglearningretrievegenerate}. 
\citet{xia2024improvingretrievalaugmentedlanguage} proposed low-impact reasoning techniques, later enhanced via contrastive reasoning by \citet{ranaldi2024elicitingcriticalreasoningretrievalaugmented}. Unlike these English-centric approaches, we focus on multilingual knowledge-intensive tasks. Complementing \cite{zhang2022makingmiraclmultilingualinformation}, we study the inference phase and enrich the work proposed by \citet{chirkova2024retrievalaugmentedgenerationmultilingualsettings,ranaldi2025multilingualretrievalaugmentedgenerationknowledgeintensive}. We propose a framework that allows the LLMs to leverage different knowledge, reason about it, and deliver argumentative explanations by using a dialectic approach. Our effort aims to improve the limitations of multilingual RAG, bias towards language, information disparity \cite{sharma2024fauxpolyglotstudyinformation} or conflicting knowledge \cite{li-etal-2024-land}.

\section{Conclusion}

RAG has demonstrated its potential to improve LLM performances in knowledge-intensive tasks; however, a major limitation lies in handling heterogeneous retrieved, especially in multilingual cases. To address this, we propose \textit{Dialectic-RAG} (\DRAG) to improve retrieval-based reasoning through argumentative explanations. We show that \DRAG significantly improves multilingual retrieval-augmented inference, enhancing both in-context learning and demonstration-based instruction for smaller models. Structuring reasoning over retrieved knowledge mitigates misleading inferences and improves response consistency, reinforcing the importance of dialectic reasoning for reliable multilingual RAG applications.

\bibliography{anthology,custom}

\newpage

\appendix

\clearpage

\begin{table*}
\section{Prompting Approaches}
\label{app:app_prompting}

\begin{small}
\begin{tcolorbox}[colback=white,colframe=customisedblue,title=Baseline Prompt Template (no RAG), rounded corners=south, rounded corners=north]

\begin{tcolorbox}[colback=white, colframe=black, rounded corners=south, rounded corners=north]
\textbf{\#Role} \\
\textit{Please answer the question by following the provided instructions.}\\
\end{tcolorbox}

\begin{tcolorbox}[colback=white, colframe=black, rounded corners=south, rounded corners=north]
\textbf{\#Instructions:}\\
\textit{Answer the question as clearly as possible based on your knowledge following the format } “\textbf{\#Answer:}” \uline{\textit{Note: answer in the query language}}. \\
\end{tcolorbox}

\begin{tcolorbox}[colback=white, colframe=black, rounded corners=south, rounded corners=north]
\textbf{\#Question:}\\
\{\texttt{\textbf{question}}\}
\end{tcolorbox}
\end{tcolorbox}
\end{small}

\caption{Baseline prompting template.}
\label{tab:baseline_prompt}

\hspace{1cm}

\begin{small}
\begin{tcolorbox}[colback=white,colframe=customisedblue,title=Baseline RAG Prompt Template, rounded corners=south, rounded corners=north]

\begin{tcolorbox}[colback=white, colframe=black, rounded corners=south, rounded corners=north]
\textbf{\#Role} \\
\textit{Please answer the question by following the provided instructions.}
\end{tcolorbox}

\begin{tcolorbox}[colback=white, colframe=black, rounded corners=south, rounded corners=north]
\textbf{\#Instructions:}\\
\textit{Answer the question as clearly as possible using the provided \textbf{Reference Evidence} and follow the format} “\textbf{\#Answer:}” \uline{\textit{Note: answer in the query language}}. \\
\end{tcolorbox}

\begin{tcolorbox}[colback=white, colframe=black, rounded corners=south, rounded corners=north]
\textbf{\#Reference Evidence:}\\
 \hspace{1cm}  \textbf{[1]} \{\texttt{\textbf{Document}}$_1$\} \\
 \hspace{1cm}  \textbf{[2]} \{\texttt{\textbf{Document}}$_2$\} \\
  \hspace{1cm}  \textbf{[3]} \{\texttt{\textbf{Document}}$_3$\} \\
   \hspace{1cm}  \textbf{[4]} \{\texttt{\textbf{Document}}$_4$\} \\
  \hspace{1cm}  \textbf{[5]} \{\texttt{\textbf{Document}}$_5$\} 

\end{tcolorbox}

\begin{tcolorbox}[colback=white, colframe=black, rounded corners=south, rounded corners=north]
\textbf{\#Question:}\\
\{\texttt{\textbf{question}}\}
\end{tcolorbox}

\end{tcolorbox}
\end{small}

\caption{RAG prompting example.}
\label{tab:baseline_RAG_prompt}

\end{table*}

\begin{table*}
\section{\DRAG prompting Template}
\label{app:DRAG-prompt-annotation}

\begin{small}

\begin{tcolorbox}[colback=white,colframe=customisedblue,title=\DRAG Prompt, rounded corners=south, rounded corners=north]

\begin{tcolorbox}[colback=white, colframe=black, rounded corners=south, rounded corners=north]
\textbf{\#Role} \\
You are helpful assistant. Please answer the question by following the provided instructions.
\end{tcolorbox}

\begin{tcolorbox}[colback=white, colframe=black, rounded corners=south, rounded corners=north]
\textbf{\#Requirements:}\\
\textit{Answer the question as clearly as possible using the provided \textbf{\#Reference Evidence} and follow the \textbf{\#Instructions}}. 
\end{tcolorbox}

\begin{tcolorbox}[colback=white, colframe=black, rounded corners=south, rounded corners=north]
\textbf{\#Reference Evidence} \\
 \hspace{1cm}  \textbf{[1]} \{\texttt{\textbf{Document}}$_1$\} \\
 \hspace{1cm}  \textbf{[2]} \{\texttt{\textbf{Document}}$_2$\} \\
  \hspace{1cm}  \textbf{[3]} \{\texttt{\textbf{Document}}$_3$\} \\
   \hspace{1cm}  \textbf{[4]} \{\texttt{\textbf{Document}}$_4$\} \\
  \hspace{1cm}  \textbf{[5]} \{\texttt{\textbf{Document}}$_5$\} 

\end{tcolorbox}

\begin{tcolorbox}[colback=white, colframe=black, rounded corners=south, rounded corners=north]
\textbf{\#Instructions} 

\begin{adjustwidth}{0.5cm}{0cm} 

    \textbf{1)} Consider the provided documents labelled “\textbf{\#Reference Evidence}”, identify and understand the main points. Follow the directions in detail and use only the information in the documents, exemplifying which points are most relevant for answering the question \textbf{\#Question}.\\
    \textit{Note: Ensure all documents are considered and provide a precise and well-structured response using English as the shared language.} Name this passage “\textbf{\#Extraction:}”. \\ 
    
    \textbf{2)} For each document, extract the most relevant information for answering the \textbf{\#Question} discussing whether they are actually \textcolor{blue}{relevant} or \textcolor{red}{irrelevant}. \\
    \textit{To ensure clarity, include the exact passages from each supporting document and reference their document numbers. Organise your argumentation as follows:" Document [1] claims [specific argument], whereas passage [4] claims...}. Name this passage as “\textbf{\#Explaination:”}.  \\ 

    \textbf{3)} Please consider the step \textbf{2)} in detail, ensure they are correct. Then, provide a single \textcolor{blue}{\uline{argumentative explanation}}
    that considers the passages and their supporting motivations from a \textit{neutral} perspective, as concern argumentative passages. \\
    \textit{Note: To enhance clarity, present your detailed explanation under the heading “\textbf{\#Dialectic Argumentation:}”}
    \\

    \textbf{4)} Finally, to facilitate the final evaluation, deliver a short-form answer by labelling it as “\textbf{\#Answer:}” \\
    \uline{\textit{Note: answer in the query language}}.

\end{adjustwidth}
    
\end{tcolorbox}

% Question
\begin{tcolorbox}[colback=white, colframe=black, rounded corners=south, rounded corners=north]
\textbf{\#Question} \\
\{\texttt{\textbf{question}}\}
\end{tcolorbox}

\end{tcolorbox}
\end{small}
\caption{The Dialectic RAG (\DRAG) framework instructs the model to deliver multi-step reasoning paths that lead the models to solve the task by explaining the perspectives that have emerged.}
\label{tab:annotation_prompt_D-RAG}

\end{table*}

\begin{table}
\section{Data Composition}
\label{app:info_dataset}
In our experiments, we use three knowledge-intensive question-answering task: \textit{(i)} MLQA \cite{lewis-etal-2020-mlqa}, \textit{(ii)} MKQA \cite{longpre2021mkqalinguisticallydiversebenchmark} and \textit{(iii)} XOR-TyDi QA \cite{asai-etal-2021-xor} as they best represent multilingual open-ended question-answering tasks. MLQA is manually translated from SQuAD v1.1 \cite{rajpurkar-etal-2016-squad}, MKQA and XOR-TyDi QA are machine translated and manually controlled by Natural Questions \cite{kwiatkowski-etal-2019-natural} and TyDi QA \cite{clark-etal-2020-tydi}, respectively. 

We use test sets in the languages in Table \ref{tab:multilingual_datasets}. For each language, we used the same questions and, consequently, the same number of questions to avoid any imbalance in double-checking by retrieving the corresponding ids. Details on the number of instances are in Table \ref{tab:num_instrances_test}.
In addition, since the experimental setting of our work requires a subset of examples to conduct the annotation phase (\S \ref{sec:DRAG_Application}), we used instances defined in Table \ref{tab:num_instrances_train} (not present in the evaluation set) and annotated them as described in Appendix \ref{app:annotation_phase_details}.

\hspace{2em}

\section{Data Annotation}
\label{app:annotation_phase_details}
We use \DRAG annotations to fine-tune smaller models to leverage knowledge-intensive tasks using retrieved documents (\S \ref{sec:tuning}). To ensure the quality of the annotations firstly, we use an exact-match as the first filter then we use GPT-4o-mini as annotator. HThen, after ensuring that the final answer matches the target, we systematically instruct the GPT-4o-mini using the \DRAG (Table \ref{tab:annotation_prompt_D-RAG}). This double-check assess the accuracy of the outcomes delivered.  
Hence, we prompt the model as follows:

\begin{small}
\begin{tcolorbox}[
    colback=gray!20,
    colframe=gray!75!black,
    colbacktitle=gray!90!white,
    fonttitle=\bfseries,
    width=\columnwidth,
    boxrule=0.5pt,
    arc=4pt,
    auto outer arc,
]
\small
\textbf{\#Role:} \\
You are an experienced expert skilled in answering complex problems through logical reasoning and structured
analysis. \\
\textbf{\#Task:} \\
Given the following sentences, you are a decision maker who decides whether the ‘Response’ provides the ‘Target’ as the final outcome and follows the given ‘Instructions‘.
If the output doesn't align with the target answer and doesn't not follow the instructions, respond with '0', whereas if it's correct, then respond with '1'. \textit{Please, ensure that all criteria are complied with the requests and do not provide any other answer beyond `0' or `1'.} \\
\textbf{\#Senteces:} \\
\#Response: \{model\_result\} \\
\#Target: \{target\_answer\}. \\
\#Instructions: \{D-RAG\_template\}. 
\end{tcolorbox}
\end{small}

\end{table}

\begin{table}

\section{Splitting Informations}
\label{sec:splitting_data}

As described in \S \ref{sec:annotation_strategy} and detailed in Appendix \ref{app:info_dataset}, we conducted an evaluation phase on equally distributed portions of the data on the analysed languages shown in Table \ref{tab:num_instrances_test}. In addition, we annotated a set of samples (Table \ref{tab:num_instrances_train}) equally distributed among the languages in Table \ref{tab:lan_training}. The annotation data were filtered separately, and although some questions are repeated for different languages (by task and dataset construction), the arguments are different because the documents retrieved are different.

\vspace{0.2cm}

\textbf{Testing Sets}\\

\begin{small}
\begin{tabular}{l|cccc}
\textbf{Dataset} & \textbf{\# per lang} & \textbf{\# per lang} & \textbf{\#Tot.} & \textbf{\#Tot.} \\ 
 & \textbf{available} & \textbf{used} & \textbf{used} & \textbf{ablation} \\ 

\hline
\hline
MLQA & 1.5$k$ & 0.8$k$ & 7.2$k$ & 1.8$k$ \\
MKQA & 2$k$ & 1.0$k$ & 6.0$k$ & 1.0$k$ \\
XOR-TyDi & 0.6$k$ & 0.4$k$ & 2.4$k$ & 0.6$k$ \\
\end{tabular}
\caption{Number (\#) of instances for evaluation (test/ablation) phases which are equally distributed among the languages in Table \ref{tab:multilingual_datasets}. ($k$ denotes 1000 instances)}
\label{tab:num_instrances_test}
\end{small}

\vspace{0.5cm}

\textbf{Training Sets}\\

\begin{small}
\begin{tabular}{l|ccc}
\textbf{Dataset} & \textbf{\#example} & \textbf{\#example} & \textbf{\#Total} \\ 
 &  & \textbf{correct} & \textbf{used} \\ 
\hline
MLQA & 3500 & 1920 & 1920 \\
MKQA & 2000 & 1128 & 920 \\
XOR-TyDi  & 800 & 556 & 200 \\
\hline
Total & 6.3$k$ & 3.6$k$ & 3.02$k$ \\
\end{tabular}
\caption{Number of datasets used for evaluation phases which are equally distributed among the languages in Table \ref{tab:multilingual_datasets}. ($k$ denotes 1000 instances)}
\label{tab:num_instrances_train}
\end{small}

\vspace{0.5cm}

\textbf{Language used for training}\\

\begin{small}
\begin{tabular}{p{2.2cm} p{5.5cm}} 
\textbf{Dataset}    & \textbf{Languages}  \\ \hline

\textbf{MKQA}      
        & English, Spanish, German, Russian, Chinese, Finnish, Arabic    \\ 
        \hline
\textbf{MLQA}     &  
        English, Chinese, Arabic, German, Spanish \\ 
\hline
\textbf{XORTyDi QA} &
        English, Chinese, Arabic, Finnish  \\ 
\hline
\end{tabular}
\caption{Languages annotation.}
\label{tab:lan_training}
\end{small}

\section{Models Version}
\label{app:model_versions}

\small
\begin{center}
\begin{tabular}{lp{5.5cm}}
\textbf{Model} & \textbf{Version}  \\ 

\hline
\hline
GPT-4o & OpenAI API (gpt-4-o)  \\
\hline
%Command-R & CohereForAI/c4ai-command-r-v01  \\
%\hline
Llama3-70B   &  meta-llama/Meta-Llama-3-70B-Instruct \\
Llama3-8B   &  meta-llama/Meta-Llama-3-8B-Instruct \\
Llama3-1B   &  meta-llama/Meta-Llama-3.2-1B-Instruct \\
\end{tabular}
\end{center}

\caption{Models versions, found on huggingface.co. We used the configurations described in \S \ref{sec:exp_set} in the repositories for each model *(access verified on 25 Jan 2025).}
\label{tab:versions_models}
\end{table}

\begin{table}[]

\section{Difference between High- and Low-resource Languages}
\label{app:low_high_resource_lan}
In this work, we define the differences between high-resource (HR) and low-resource (LR) using the consideration already taken in previous works \cite{ranaldi-pucci-2023-english,ranaldi-etal-2024-empowering-multi}.
We report two tables: Table \ref{tab:language_distribution} reports the language distribution of CommonCrawl, and Table \ref{tab:language_distribution_wikimedia_dump} the number of documents in the Wikipedia dump used in our work (\S \ref{sec:exp_set}).

\begin{center}
\small
\begin{tabular}{|l|c|}
\hline
\textbf{Language} & \textbf{Percentage} \\ \hline
English (en)      & 46.3\%              \\ \hline
Russian (ru)      & 6.0\%               \\ \hline
German (de)       & 5.4\%               \\ \hline
Chinese (zh)      & 5.3\%               \\ \hline
French (fr)       & 4.4\%               \\ \hline
Japanese (ja)     & 4.3\%               \\ \hline
Spanish (es)      & 4.2\%               \\ \hline
Other             & 23.1\%              \\ \hline
\end{tabular}
\caption{Language distribution of CommonCrawl \cite{commoncrawl2021}.}
\label{tab:language_distribution}
\end{center}

\section{Documents in Wikimedia\_Dump}
\begin{center}
\small
\begin{tabular}{|l|l|}
\hline

\textbf{Language} & \textbf{Percentage} \\ \hline
English (en)      & 41,488k              \\ 
Russian (ru)      & 13,784k               \\ 
German (de)       & 20,772k               \\ 
Chinese (zh)      & 7,875k               \\ 
Italian  (it)         & 10,462k   \\ 
French (fr)       & 17,813k               \\ 
Japanese (ja)     & 6,626k               \\ 
Spanish (es)      & 12,865k               \\ 
Portuguese (pt)              & 5,637k               \\
Bengali (bn)      & 767k               \\ 
Finnish (fn)          & 272k          \\
Arabic (ar)          &   1,050k      \\
Thai (th)          &      876k   \\
Vietnamese (vi)      & 2,067k \\
Telogu (te)       &  124k \\ \hline

\end{tabular}
\caption{Language distribution of Wikimedia Dump introduced in \S \ref{sec:exp_set}.}
\label{tab:language_distribution_wikimedia_dump}
\end{center}

\end{table}

\begin{table}[t]

\section{Retrieval Details}
\label{app:retrieval}
\paragraph{Retrieval}
We use Cohere as the retrieval system and Wikimedia\_dump as the knowledge base $\mathcal{K}$ for all experiments. We use $\mathcal{K}$ provided by Cohere \textit{wikipedia-2023-11-embed-multilingual-v3} (available on \href{Cohere/wikipedia-2023-11-embed-multilingual-v3}{huggingface}). They provide individual documents embedded with multilingual embedding model \textit{Cohere\_Embed\_V3} (in Table \ref{tab:language_distribution_wikimedia_dump} are reported the dump composition). For each question in the evaluation data, we retrieve 10 relevant documents and then filter the top-5 most relevant ones as done in the related \href{Cohere/wikipedia-2023-11-embed-multilingual-v3}{repository} (dot score between query embedding and document embeddings).
\end{table}

\begin{table}[h]
\section{Ablation Argumentation Language}
\label{app:ablation_language}
\DRAG is instructed to use an English argumentation (see Table \ref{tab:annotation_prompt_D-RAG}). In this experiment, we instruct the model to operate in Chinese, Arabic and German and report the differences with the original \DRAG, which is in English.

\hspace{0.2cm}
\begin{small}
  \begin{tabularx}{0.48\textwidth}{p{1.8cm}<{\raggedright}p{2cm}<{\centering}p{1cm}<{\centering}p{1cm}<{\centering}}
    \toprule
    \multirow{1}{*}{ \textbf{Models}+\DRAG} & \textbf{$\Delta$\textsc{De}} & \textbf{$\Delta$\textsc{Zh}} & \textbf{$\Delta$\textsc{Ar}} \\
%    & \textbf{(\DRAG)} &  & \\
\midrule
    
GPT-4o      & -2.4 & -6.3 & -8.6 \\
Llama3-70B  & -6.8 & -9.5 &  -12.6 \\
Llama3-8B  & -8.1 & -9.3 &  -14.6 \\
Llama3-1B  & \textbf{-12.8} & \textbf{-16.6} &  \textbf{-18.4} \\

  \bottomrule
\end{tabularx}
\end{small}
\caption{Ablation on argumentation language impacts on \DRAG using MKQAs' ablation set.}
  \label{tab:ablation_query_language}

\section{Ablation Output Analysis}
\label{app:ablation_outputs}
To control the quality of the generations, we defined two different metrics: Instruction Following (IF) and Correct Language (CL). The role of IF is to investigate whether the models followed the instructions given in the prompt. The role of CL, on the other hand, is to analyse whether the language of the final response is the same as that of the query (note that this requirement was well defined in the prompt.
In order to have a robust result, we conducted these two analyses using GPT-4o-mini as an instructed evaluator, using the prompt in Appendix \ref{app:annotation_phase_details} and avoiding the target part in the case of IF. We computed the CL using OpenLID framework \cite{burchell-etal-2023-open}.
For both values, we reported the percentage of correctness (accuracy).

\end{table}

\begin{table}[h]
\section{Ablation number of Steps}
\label{app:ablation_steps}
\DRAG operates via a single instruction. To observe the impact of instruction splitting on the final performances, we apply the same prompt shown in Table \ref{tab:annotation_prompt_D-RAG} by giving the model one step at a time. 

\hspace{0.2cm}
\small
\center
  \begin{tabularx}{0.46\textwidth}{p{1.8cm}<{\centering}|p{1cm}<{\centering}p{1cm}<{\centering}p{1cm}<{\centering}}
   \hline
    \multirow{1}{*}{ \textbf{Models} } & \multicolumn{1}{c}{\textbf{MKQA}} & \multicolumn{1}{c}{\textbf{MLQA}} & \multicolumn{1}{c}{\textbf{XoR TyDi}} \\
    \hline
    \multicolumn{4}{c}{\textbf{GPT-4o}} \\  
    \hline
Single Step  & \textbf{68.6} & 65.8 & 61.3 \\
4 Steps  & 68.4 & 66.9 & 63.0 \\
\hline
    \multicolumn{4}{c}{\textbf{Llama3-70B}} \\  
    \hline
Single Step  & 67.0 & 62.9 & 56.2  \\
4 Steps  & \textbf{67.5} & \textbf{63.4} & \textbf{56.0} \\
\hline
    \multicolumn{4}{c}{\textbf{Llama3-8B tuned via \DRAG }} \\  
\hline
Single Step  & 62.4 & 59.8 & 52.1 \\
4 Steps  & \textbf{63.5} & \textbf{60.9} & \textbf{53.6} \\
\hline
    \multicolumn{4}{c}{\textbf{Llama3-8B tuned via \DRAG }} \\  
   \hline
Single Step  & 55.9 & 53.2 & 46.4  \\
4 Steps  & \textbf{57.4} & \textbf{55.3} & \textbf{48.9} \\

  \bottomrule
\end{tabularx}
\caption{\DRAG using Single Step prompting (traditional approach) and breaking the steps into single phases on ablation set of proposed QA tasks.}
  \label{tab:ablation_num_ICL}
\end{table}

\begin{table*}[]

\section{Proposed Task}
\label{app:languages}
\begin{center}
    
\small
\begin{tabular}{p{2cm} p{10cm} p{1.5cm}} 
\hline
\textbf{Dataset}    & \textbf{Languages} & \textbf{\#Languages} \\ \hline

\textbf{MKQA}      
        & \texttt{English, Spanish, German, Russian, Chinese, Finnish, Arabic, \textbf{Italian}, \textbf{Korean}} 
        & 9 \\ 
\hdashline

\textbf{MLQA}       
        & \texttt{English, Chinese, Arabic, German, Spanish, \textbf{Hindi}} 
        & 6 \\ 
\hdashline

\textbf{XORTyDi QA} 
        & \texttt{English, Chinese, Arabic, Finnish, \textbf{Korean}, \textbf{Telugu}} 
        & 6 \\ 
%\hdashline

\hline
\end{tabular}
\end{center}
\caption{Languages present in datasets used in this work. *In \textbf{bold}, the languages are used only for evaluation as described in Appendix \ref{app:info_dataset}.}
\label{tab:multilingual_datasets}

\section{Experiment on \textsc{BorderLines}}
\label{app:exps_borderlines}
To investigate the impact of our \DRAG in real contexts, we used examples from the \textsc{BorderLines} \cite{li-etal-2024-land}. This resource has questions concerning disputed territories between two nations that bureaucratically belong to a specific country. The questions have the form \texttt{Is \textbf{Place P} a territory of \textbf{A) Country X} or \textbf{B) Country Y}?}.
These questions are in English, language \textbf{\texttt{X}} and \textbf{\texttt{Y}} (are the languages spoken in the countries). Finally, a target or controller value indicates the country that controls the \texttt{\textbf{P}}. (in some cases, there are no defined places that we do not consider in our analysis)

To study the consistency and dialectic capabilities of our \DRAG, we selected a small set consisting of 60 instances (20 questions in English, l20 language \textbf{\texttt{X}} and 20 \textbf{\texttt{Y}}). We then conducted a retrieval phase and prompt GPT-4o with the questions in the specific languages and English using the prompts defined in Appendices \ref{app:app_prompting} and \ref{app:DRAG-prompt-annotation}.
Then, we set the controller as \textbf{\texttt{X}} we estimated the percentage of times the answer provided by the model prompted in English matched with the controller (denoted as \textbf{\%Agreement English}, and the percentage when the models prompted via queries in three languages matches among them and with the controller.
\end{table*}

\begin{table*}[h!]
\section{\DRAG and GPT-4o on \textsc{BorderLines}}
\label{app:example_Knowledge_Conflicts}
\centering
\small
\begin{tabular}{p{16cm}}
$\bullet$ $\mathcal{Q}_{En}$: \textit{Is Sixty-Four Villages East of the River a territory of A) Russia or B) Republic of China?} \\
$\bullet$ $\mathcal{Q}_{Ru}$:  \begin{tiny}
    \foreignlanguage{russian}{Являются ли «Шестьдесят четыре деревни к востоку от реки» территорией А) России Б) Республики Китай?} \end{tiny} \\ 
$\bullet$ $\mathcal{Q}_{Zh}$: \begin{CJK}{UTF8}{gbsn}河东六十四村是A）俄罗斯还是B）中华民国的领土？\end{CJK} \\

$\bullet$ \textbf{Target:} A) \textit{Russia} (as reported in \textsc{BorderLines} \cite{li-etal-2024-land})
 \\
\midrule
\textbf{Baseline}\\

\textbf{Answer to \textbf{$\mathcal{Q}_{En}$:}} I couldn't find any information about 'Sixty-Four Villages East of the River'.\\

\textbf{Answer to \textbf{$\mathcal{Q}_{Zh}$:}} \begin{CJK}{UTF8}{gbsn}河東六十四鄉是 B) 中華民國的領土。\end{CJK}  \\

\textbf{Answer to \textbf{$\mathcal{Q}_{Ru}$:}} \foreignlanguage{russian}{Правильный ответ: A) Россия.}\\

\midrule
\textbf{Retrieved documents}\\

[1] The Sixty-Four Villages East of the River were inhabited villages by Manchu, Daur, and Han, located on the left (north) bank of the Amur River opposite Heihe and the Blagoveshchensk. Among Russian historians, the district occupied by the villages referred to as Zazeysky [...] \textcolor{red}{(from \textbf{W}$_{En}$)}  \\

[2] The Treaty of Aigun of 1858, signed by the Chinese Qing Dynasty and the Russian Empire, ceded the lands north of the Amur to Russia.
Manchu residents who lived on the left bank of the Amur received the right to stay and live on Russian territory. During the Boxer Rebellion of 1900, Russia sent its troops to occupy 64 villages and forced the people to cross the Amur River and settle in China. Currently, most of this territory belongs to the Blagoveshchensky and Tambov districts of the Amur region. \textcolor{red}{(from \textbf{W}$_{Ru}$)}\\

[3] The ‘Treaty of Peking’ signed between the Qing Dynasty and the Russian Empire stated: ‘The land mentioned above is open land. Wherever the Chinese live, or the Chinese occupy the fishing and hunting land, Russia is not allowed to occupy it. The Chinese may still occupy it by fishing and hunting as usual. The Boxer Rebellion took place in 1900, and the Qing government had no time to consider the situation in the northeast. Russia used the destruction of the Dongqing railway by the Boxer Rebellion as an excuse to send troops [...]  \textcolor{red}{(from \textbf{W}$_{Zh}$)}\\

[4] The 1900 Amur anti-Chinese pogroms were a series of ethnic killings (pogroms) and reprisals undertaken by the Russian Empire against subjects of the Qing dynasty of various ethnicities. They took place in the Sixty-Four Villages East of the River in the Amur region during the same time as the Boxer Rebellion. The events ultimately resulted in thousands of deaths, the loss of residency for Chinese subjects living in the villages, and increased Russian control over the region.[...] \textcolor{red}{(from \textbf{W}$_{En}$)}\\

[5] The Boxer Rebellion, also known as the Boxer Uprising, was an anti-foreign, imperialist, and Christian uprising in North China between 1899-1901, towards the end of the Qing dynasty, by the Society of Righteous and [...] \textcolor{red}{(from \textbf{W}$_{En}$)}\\

\midrule
\textbf{RAG} \\
\textbf{Answer to \textbf{$\mathcal{Q}_{En}$:}} The Sixty-Four Villages East of the River The villages are currently a territory of \textbf{A) Russia}.\\
\textbf{Answer to \textbf{$\mathcal{Q}_{Zh}$:}} \begin{CJK}{UTF8}{gbsn}河东六十四村历史上属于清朝（中华民国前身）领土，根据《北京条约》规定，俄罗斯不得占据中国居民捕鱼狩猎的土地。[...] 尽管该地区现由俄罗斯实际控制，但从历史和法理来看，应属中华民国领土。

答案：B）中华民国 \end{CJK}  \textcolor{red}{(\textbf{B) China)}}\\ 

\textbf{Answer to \textbf{$\mathcal{Q}_{Ru}$:}} \foreignlanguage{russian}{Шестьдесят четыре деревни к востоку от реки стали частью России по Айгунскому (1858) и Пекинскому (1860) договорам. В 1900 году события восстания ихэтуаней усилили российский контроль. Сегодня эта территория входит в состав Амурской области России. 
Ответ: \textbf{A) Россия.}} \textcolor{red}{\textbf{(A) Russia)}}\\

\midrule
\textbf{\DRAG} \\

\textbf{\#Extraction:}\\
The Sixty-Four Villages East of the River were historically 
inhabited by Manchu, Daur, and Han peoples, located on the bank of the Amur and east of the Zeya (Ref [1],[3]). These villages became a contested region following treaties and military actions between China and Russia. Key points include:

1. Territorial Cession via Treaties: [...] (Ref [2], [3]).\\

2. Russian Control Post-1900 [...] (References [3], [4]).

3. Modern Status: region currently controlled by Russia (References [2], [3]).\\ 

\textbf{\#Explaination:}\\
Doc [1]: Describes the geographic and ethnic composition of the villages and emphasizes their location and their original habitation by Chinese citizens. Relevant information as it establishes the historical context of the region.

Doc [2]: States that the Aigun Treaty ceded lands north of the Amur to Russia but allowed Chinese residents to remain and mentions Russian actions during the Boxer Rebellion. Relevant Information as it explains the historical transition.

Doc [3]: Adds detail about the Beijing Treaty, confirming Chinese residents retained rights to the land even after it was ceded. It highlights that Russian forces took advantage of the Qing’s preoccupation with the Boxer Rebellion to fully seize control. Relevant Information to the question.

Doc [4]: Focuses on the anti-Chinese pogroms during the rebellion, detailing Russian motivations for taking control of the region. Relevant Information as it corroborates the transition of control through violence and expulsion.

Doc [5]: Provides context on the Boxer Rebellion but is less directly relevant to the territorial question. It helps explain the instability that allowed Russian actions. Partially relevant information.

\textbf{\#Dialectic Argumentation:}\\
The Sixty-Four Villages East of the River historically belonged to China, as evidenced by treaties such as the Aigun and Beijing agreements, which ceded land to Russia but retained Chinese habitation rights. While the land was historically connected to Qing China, its effective governance shifted to Russia after 1900, and it remains under Russian control today as part of Amur Oblast. Historically, the territory transitioned from Chinese habitation to Russian control due to military force and treaty reinterpretations. This aligns with contemporary boundaries where the area is recognized as Russian.\\
\textbf{\#Answer:}\\
\begin{CJK}{UTF8}{gbsn}A) 俄罗斯 \end{CJK}\\  \foreignlanguage{russian}{\textbf{A) Россия.}} \\
\textbf{A) Russia}\\
\hline
\end{tabular}
\caption{\DRAG inference compared to other approaches on \textsc{BorderLines} \cite{li-etal-2024-land}.}
\end{table*}

\begin{table*}[h!]
\section{Example Outputs MKQA}
\label{app:example}
\centering
\small
\begin{tabular}{p{16cm}}
\textbf{Question:} \textit{¿quién escribió variaciones de Campanita del lugar?} \\
\textbf{Target:} \textit{Wolfgang Amadeus Mozart.}
 \\
\midrule
\textbf{Llama-3-8B (baseline no RAG)} 
Respuesta: Ludwig van Beethoven. \\
\midrule
\textbf{Llama-3-8B (RAG)} \\
The answer is \textbf{Wolfgang Amadeus Mozart.} \\
\midrule
\textbf{\DRAG} \\
\\

\textbf{\#Extraction:}\\
The question asks who wrote variations of Campanita del lugar, which is the Spanish version of Twinkle, Twinkle, Little Star.\\
\\
\textbf{\#Explaination:}\\
Document [1] states that Wolfgang Amadeus Mozart composed Twelve Variations on "Ah vous dirai-je, Maman", a French folk song, which is the same melody used for Campanita del lugar.\\

Document [2] mentions that the melody of Campanita del lugar is based on Mozart's music, but it does not explicitly state that Mozart wrote variations of Campanita del lugar.\\

Document [5] explains that "Twinkle, Twinkle, Little Star" is sung to the tune of the French melody "Ah! vous dirai-je, maman", which was first published in 1761 and later arranged by several composers, including Mozart with Twelve Variations on "Ah vous dirai-je, Maman".\\

Documents [3] and [4] are irrelevant to the question.\\
\\
\textbf{\#Dialectic Argumentation:}\\
Based on the provided documents, it is clear that Mozart wrote Twelve Variations on "Ah vous dirai-je, Maman", which is the same melody used for Campanita del lugar. While Mozart did not specifically write variations of Campanita del lugar, his work on the French folk song is the foundation of the melody used for the Spanish version.\\
\\

\textbf{\#Answer:}\\

Wolfgang Amadeus Mozart escribió variaciones de la melodía utilizada para Campanita del lugar, en particular Doce variaciones sobre 'Ah vous dirai-je, Maman'.
\\

\textbf{La respuesta es Wolfgang Amadeus Mozart.}
\\
\\
\hline
\end{tabular}
\caption{An example of \DRAG inference compared to other approaches. While the baseline answer is wrong, the RAG answer is correct but delivers an English text. \DRAG deliver the reasoning steps that supports the answer and the final answer in query language.}
\end{table*}

\end{document}